\documentclass{article}

\usepackage{times}
\usepackage{graphicx} 
\usepackage{subfigure} 

\usepackage{natbib}

\usepackage{algorithm}
\usepackage{algorithmic}

\usepackage{hyperref}

\makeatletter
\providecommand\theHALG@line{\thealgorithm.\arabic{ALG@line}}
\makeatother

\usepackage{amsfonts}
\usepackage{amsmath}
\usepackage{amssymb}
\usepackage{amsthm}
\usepackage{mathtools}
\usepackage{lmodern}

\renewcommand{\algorithmicrequire}{\textbf{Input:}}

\usepackage[textsize=footnotesize,color=green!40]{todonotes}

\usepackage{stmaryrd}

\usepackage{xcolor}
\hypersetup{
    colorlinks,
    linkcolor={red!80!black},
    citecolor={blue!80!black},
    urlcolor={blue!80!black}
}

\newcommand{\comment}[1]{}

\renewcommand{\comment}[1]{}

\newcommand{\R}{\mathbb{R}}

\renewcommand{\vec}[1]{\ensuremath{\mathbf{#1}}}
\newcommand{\vecs}[1]{\ensuremath{\mathbf{\boldsymbol{#1}}}}
\newcommand{\mat}[1]{\ensuremath{\mathbf{#1}}}
\newcommand{\mats}[1]{\ensuremath{\mathbf{\boldsymbol{#1}}}}
\newcommand{\ten}[1]{\mat{\ensuremath{\boldsymbol{\mathcal{#1}}}}}
\newcommand{\ttm}[1]{\times_{#1}}
\newcommand{\ttv}[1]{\bullet_{#1}}
\newcommand{\tenmat}[2]{\mat{#1}_{(#2)}}

\newcommand{\nsample}[1]{^{(#1)}}

\newcommand{\T}{\ten{T}}
\newcommand{\W}{\ten{W}}
\newcommand{\G}{\ten{G}}
\newcommand{\Y}{\ten{Y}}
\newcommand{\U}{\mat{U}}
\newcommand{\X}{\mat{X}}
\newcommand{\I}{\mat{I}}
\newcommand{\M}{\mat{M}}
\newcommand{\K}{\mat{K}}
\renewcommand{\v}{\vec{v}}
\newcommand{\x}{\vec{x}}
\newcommand{\tX}{\tilde{\X}}
\newcommand{\tY}{\tilde{\Y}}

\newtheorem*{theorem*}{Theorem}
\newtheorem{theorem}{Theorem}
\newtheorem{proposition}{Proposition}

\DeclareMathOperator*{\Tr}{Tr} 

\newcommand{\vectorize}[1]{\mathrm{vec}(#1)}

\DeclareMathOperator*{\rank}{rank}
\newcommand{\norm}[1]{\|#1\|}
\newcommand{\kron}{\otimes}

\newcommand{\bigo}[1]{\mathcal{O}(#1)}

\title{Higher-Order Low-Rank Regression}
\author{Guillaume Rabusseau\footnote{Contact author: \texttt{guillaume.rabusseau@lif.univ-mrs.fr}.} \qquad Hachem Kadri \\ Aix-Marseille University  }

\begin{document}
\maketitle

\begin{abstract} 

This paper proposes an efficient algorithm (HOLRR) to handle regression tasks where
the outputs have a tensor structure. We formulate the regression problem as the minimization 
of a least square criterion under a multilinear rank constraint, a difficult  non convex problem. 
HOLRR computes efficiently an approximate solution of this problem, with solid theoretical guarantees.
A kernel extension is also presented. Experiments on synthetic and real
data show that HOLRR outperforms multivariate and multilinear regression methods and
is considerably faster than existing tensor methods.

\comment{
The predictive accuracy of HOLRR is assessed
on experiments 

Extending univariate and multivariate methods to tensor-structured data is a challenging task which has recently 
received a growing interest.
In this paper, we generalize the reduced-rank regression method from vector to tensor-structured response variables. 
Reduced-rank regression allows one to capture dependencies in the response variables by enforcing the regression
matrix to be of low rank, but it is unable to capture higher-order dependencies. 
To address this limitation we propose to learn a regression tensor of low multilinear
rank. We design an efficient algorithm for which we provide theoretical approximation guarantees and a kernelized extension.
We compare our method  with other multi-output regression methods on both synthetic and real data 
where it exhibits better predictive accuracy than standard multivariate methods and is more computationally
efficient than existing tensor regression methods.}

\end{abstract} 

\section{Introduction}

Recently, there has been an increasing interest in adapting machine learning and statistical methods to tensors.
 Data with a natural tensor structure is encountered in many scientific areas
 including neuroimaging~\citep{Zhou2013}, signal processing~\citep{Cichocki2009}, spatio-temporal analysis~\citep{Bahadori2014} 
 and computer vision~\citep{Lu2013}.
Extending multivariate regression methods to tensors is one of the challenging task in this area.
Most existing works extend linear models to the multilinear setting and focus on the tensor structure 
of the input data (e.g.~\citet{signoretto2013a}). Little has been done however to investigate  learning methods 
for \emph{tensor-structured output data}. 

We consider a multilinear regression task where inputs are vectors and outputs are tensors. 
In order to leverage the tensor structure of the output data, we formulate the problem as the
minimization of a least squares criterion 
subject to a \emph{multilinear rank} constraint on the regression tensor. The rank constraint enforces 
the model to capture the low-rank structure of the outputs and to explain
the dependencies between inputs and outputs in a low-dimensional multilinear subspace. 

Unlike previous work we do not use a convex relaxation of this difficult non-convex optimization problem. Instead we  design
an efficient approximation algorithm (HOLRR) for which we are able to provide good approximation guarantees.
We also present a kernelized version of HOLRR which extends our model to the nonlinear
setting. Experiments on synthetic and real data shows that HOLRR obtains better predictive accuracy
while being computationally very competitive. We also present an image recovery experiment
which gives an illustrative insight on the effects of multilinear rank regularization.


\paragraph{Related work.}  In the context of multi-task learning, \citet{romera2013} have proposed a linear model using a 
tensor-rank penalization of a least squares criterion  to take into account the multi-modal interactions between tasks. Their
approach relies on a convex relaxation of the multlinear rank constraint using the trace norms of the matricizations. They also
propose a non-convex approach (MLMT-NC) but it is computationally very expensive.
\citet{Bahadori2014} have proposed a greedy algorithm to solve a low-rank tensor learning problem in the context of multivariate 
spatio-temporal data analysis. The linear model they assume is different from the one we propose, it is specifically 
designed for spatio-temporal data and does not fit into the general tensor-valued regression framework we consider here.

\paragraph{Paper outline.}
We provide some background on low-rank regression and tensors in Section~\ref{sec:prelim}. 
In Section~\ref{sec:algo} we introduce the tensor response regression problem,
we develop the HOLRR algorithm to tackle the minimization problem with mulitlinear rank constraint, and
we prove that this algorithm has good approximation guarantees. A kernelized version of HOLRR is
provided in Section~\ref{sec:kernel}. Finally, we assess the performances of our method through simulation
study and real data analysis in Section~\ref{sec:xp}.

\section{Preliminaries}
\label{sec:prelim}
We begin by introducing some notations.
For any integer $k$ we use $[k]$ to denote the set of integers from $1$ to $k$.
We use lower case bold letters  for vectors (e.g.\ $\vec{v} \in \R^{d_1}$),
upper case bold letters for matrices (e.g.\ $\M \in \R^{d_1 \times d_2}$) and
bold calligraphic letters for higher order tensors (e.g.\ $\T \in \R^{d_1
\times d_2 \times d_3}$).
The identity matrix will be written as $\I$.
The $i$th row (resp. column) of a matrix $\M$ will be denoted by
$\M_{i,:}$ (resp. $\M_{:,i}$). This notation is extended to
slices of a tensor in the straightforward way.
If $\vec{v} \in \R^{d_1}$ and $\vec{v}' \in \R^{d_2}$, we use $\vec{v} \kron \vec{v}' \in \R^{d_1
\cdot d_2}$ to denote the Kronecker product between vectors, and its
straightforward extension to matrices and tensors.
Given a matrix $\M \in \R^{d_1 \times d_2}$, we use $\vectorize{\M} \in \R^{d_1
\cdot d_2}$ to denote the column vector obtained by concatenating the columns of
$\M$.

\subsection{Tensors and Tucker Decomposition}
We first recall some basic definitions of tensor algebra; more details can be found
in~\citet{Kolda09}. 
A \emph{tensor} $\T\in \R^{d_1\times\cdots \times d_p}$ is a collection
of real numbers $(\T_{i_1,\cdots,i_p}\ : \ i_n\in [d_n], n\in [p])$. The
\emph{mode-$n$} fibers of $\T$ are the vectors obtained by fixing all
indices except for the $n$th one, e.g. $\T_{:,i_2,\cdots,i_p}\in\R^{d_1}$.
The \emph{$n$th mode matricization} of $\T$ is the matrix having the
mode-$n$ fibers of $\T$ for columns and is denoted by
$\tenmat{T}{n}\in \R^{d_n\times d_1\cdots d_{n-1}d_{n+1}\cdots d_p}$.
The vectorization of a tensor is defined by $\vectorize{\T}=\vectorize{\tenmat{T}{1}}$.
The \emph{inner product} between two tensors $\ten{S}$ and $\ten{T}$ (of the same size)
is defined by $\langle \ten{S},\T\rangle = \langle\vectorize{\ten{S}},\vectorize{\T}\rangle$
and the Frobenius norm is defined by $\norm{\T}^2_F = \langle \T,\T\rangle$.
In the following $\T$ always denotes a tensor of size $d_1\times\cdots \times d_p$.

\paragraph{mode-$n$ product.} The \emph{mode-$n$ matrix product} of the tensor $\T$ and a matrix
$\X\in\R^{m\times d_n}$ is a tensor  denoted by $\T\ttm{n}\X$. It is 
of size $d_1\times\cdots \times d_{n-1}\times m \times d_{n+1}\times
\cdots \times d_p$ and is defined by the relation 
$\Y = \T\ttm{n}\X \Leftrightarrow \tenmat{Y}{n} = \X\tenmat{T}{n}$.
The \emph{mode-$n$ vector product} of the tensor $\T$ and a vector
$\vec{v}\in\R^{d_n}$ is a tensor defined by $\T\ttv{n}\vec{v} = \T\ttm{n}\vec{v}^\top
\in \R^{d_1\times\cdots \times d_{n-1}\times d_{n+1}\times
\cdots \times d_p}$.
Given tensors $\ten{S}$ and $\T$ and matrices $\X,\mat{A}$ and $\mat{B}$,
it is easy to check that
$\langle \ten{T}\ttm{n}\X, \ten{S} \rangle= \langle \ten{T},\ten{S}\ttm{n}\X^\top\rangle$ and 
$(\T\ttm{n}\mat{A})\ttm{n}\mat{B} = \T\ttm{n}\mat{BA}$ (where we assumed conforming
dimensions of the tensors and matrices).

\paragraph{Multilinear rank.} The \emph{mode-$n$ rank} of $\T$ is the dimension of the space
spanned by its mode-$n$ fibers, that is $\rank_n(\T) = \rank(\tenmat{T}{n})$. 
The \emph{multilinear rank} of $\T$, denoted by $\rank(\T)$, is the tuple of mode-$n$
 ranks of $\T$:  $rank(\T) = (R_1,\cdots,R_p)$ where $R_n = \rank_n(\T)$ for $n\in[p]$.
 We will write $\rank(\T)\leq (S_1,\cdots,S_p)$ whenever 
 $\rank_1(\T)\leq S_1,\rank_2(\T)\leq S_2,\cdots,\rank_p(\T)\leq S_p$.
 
\paragraph{Tucker decomposition.} The \emph{Tucker decomposition} is a form of
higher-order principal component analysis. It decomposes a tensor $\T$ into a
core tensor $\G$ transformed by an orthogonal matrix along each mode:
\begin{equation}
\label{eq:tucker}
\T = \G \ttm{1}\U_1\ttm{2}\U_2\ttm{3}\cdots\ttm{p}\U_p\enspace,
\end{equation}
where $\G\in \R^{R_1\times R_2 \times \cdots \times R_p}$, $\U_i \in \R^{d_i\times R_i}$ for $i\in[p]$
and $\U_i^\top \U_i = \I$ for all $i\in [p]$. The number of parameters involved in a Tucker decomposition
can be considerably smaller than $d_1d_2\cdots d_p$.
We have the following identities when matricizing and
vectorizing a Tucker decomposition
\begin{align}
\hspace{-0.1cm}\tenmat{T}{n} &= \U_n \tenmat{G}{n}(\U_p\kron\cdots\kron\U_{n+1}\kron\U_{n-1}\kron\cdots\kron\U_1)^\top\nonumber \\
&\vectorize{\T} = (\U_p\kron \U_{p-1}\kron \cdots\kron\U_1)\vectorize{\G}\enspace.\label{eq:vecTucker}
\end{align}

It is well known that $\T$ admits the Tucker 
decomposition~(\ref{eq:tucker}) if and only if  $\rank(\T)\leq  (R_1,\cdots,R_p)$ (see e.g.~\citet{Kolda09}).
Finding an exact Tucker decomposition can be done using the higher-order SVD algorithm (HOSVD) introduced
by \citet{de2000multilinear}.
Although finding the best approximation of multilinear rank $(R_1,\cdots,R_p)$ of a tensor $\T$ is a difficult
problem,  the truncated HOSVD algorithm provides good approximation guarantees 
and often performs well in practice.

\subsection{Low-Rank Regression}
\label{sec:LRR}
Multivariate regression is the task of recovering a  function $f: \R^d \to \R^p$ from a set of input-output 
pairs $\{(\vec{x}^{(n)}, \vec{y}^{(n)})\}_{n=1}^N$, where the outputs are sampled from the model with an additive 
noise $\vec{y} = f(\vec{x}) + \vecs{\varepsilon}$, where $\vecs{\varepsilon}$ is the error term. 
To solve this problem, the \emph{ordinary least squares}~(OLS) approach assumes a linear dependence between
 input and output data and boils down to finding a matrix $\vec{W} \in \R^{d \times p}$ that minimizes the 
 squared error $\| \vec{XW} - \vec{Y}\|^2_F$, where $\vec{X} \in \R^{N\times d}$ and $\vec{Y} \in \R^{N\times p}$  
 denote the input and  the output matrices. To prevent overfitting and to avoid
 numerical instabilities a ridge regularization term (i.e. $\gamma\norm{\mat{W}}^2_F$) is often added to
 the  objective function, leading to the \emph{regularized least squares}~(RLS) method.
 The main drawback of this method is that it does not take into account the dependencies between the components 
 of the response. It performs poorly when the outputs are correlated and the true dimension of the response is less than $p$.
  Indeed, it is easy to see that the OLS/RLS approach in the multivariate setting is equivalent to performing 
  $p$ linear regressions for each scalar output $\{\vec{y}_j\}_{j=1}^p$ independently.

\emph{Low-rank regression} (or reduced-rank regression) addresses this issue by solving the following rank penalized problem
\begin{equation}
\label{eq:matrixRRR}
 \min_{\mat{W}\in \R^{d\times p}} \| \vec{XW} - \vec{Y}\|^2_F + \gamma\norm{\mat{W}}^2_F 
  \mbox{ s.t. } \rank(\vec{W}) \leq R\enspace,
 \end{equation}
for a given integer $R$. 

Suppose that $f(\x)=\mat{W}\x$. The constraint $rank(\vec{W}) = R$ implies linear restrictions on $\vec{W}$, i.e. there must exist $p-R$ linearly 
independent vectors $\v_1, \cdots, \v_{p-R} \in \R^p$ such that $\vec{W}\v_i = \vec{0}$, 
which, in turn, implies that 
$\vec{y}^\top \v_i =  \vecs{\varepsilon}^\top \v_i$ 
for each $i\in[p-R]$. Intuitively this means that the linear subspace generated by the $\v_i$'s only contains noise. 
Another way to see this is to write $\vec{W} = \vec{AB}$ with $\vec{A}\in \R^{d\times R}$ and $\vec{B} \in \R^{R\times p}$, 
which implies that the relation between $\vec{x}$ and $\vec{y}$ is explained in an $R$-dimensional latent space. 

The rank constraint in~(\ref{eq:matrixRRR})  was first proposed in~\citep{Anderson51}, 
whereas the term \emph{reduced-rank regression} was introduced in~\citep{Izenman75}.
Adding a  ridge regularization to the rank penalized problem was proposed in~\cite{Mukherjee2011}. 
In the rest of the paper we will refer to this approach as low-rank regression (LRR).
The solution of  minimization
problem~(\ref{eq:matrixRRR}) is given by projecting the RLS solution onto the space spanned by the top $R$
eigenvectors of $\mat{Y}^\top\mat{P}\mat{Y}$ where $\mat{P}$ is the orthogonal projection 
matrix onto the column space of $\X$, that is $\mat{W}_{LRR} = \mat{W}_{RLS}\mats{\Pi}$ where
$\mats{\Pi}$ is the matrix of the aforementioned projection.  
For more description and discussion of reduced-rank regression, we refer
the reader to the books of~\citet{Reinsel1998} and~\citet{Izenman08}.

%


\section{Low-Rank Regression for Tensor-Valued Functions}
\label{sec:algo}
\subsection{Problem Formulation}

We consider a multivariate regression task where the response has a tensor structure. 
Let  $f: \R^{d_0}\to\R^{d_1\times d_2\times\cdots\times d_p}$ be the function
we want to learn from a sample of input-output data $\{(\vec{x}\nsample{n}, \ten{Y}\nsample{n})\}_{n=1}^N$ 
drawn from the model  $\ten{Y} = f(\vec{x}) + \ten{E}$, where $\ten{E}$ is an error term.
We assume that the function $f$ is linear,
that is $f(\vec{x}) = \ten{W} \ttv{1} \vec{x}$ for some regression tensor 
$\ten{W} \in  \R^{d_0\times d_1 \times \cdots \times d_p}$. 
Note that the vectorization of this relation leads to $\vectorize{f(\vec{x})} = \tenmat{W}{1}^\top \vec{x}$
showing that this model is equivalent to the standard multivariate linear model.

\paragraph{Vectorization of the outputs.}
One way to tackle this linear regression task for tensor responses would be to vectorize 
each output sample and  to perform a standard multivariate low-rank regression 
on the data $\{(\vec{x}\nsample{n}, \vectorize{\Y\nsample{n}})\}_{n=1}^N\subset \R^{d_0}\times\R^{d_1\cdots d_p}$. 
%
A major drawback of this approach is that the tensor structure of the output is lost in the vectorization step. 
The low-rank model tries to capture linear dependencies between components of the output 
but it ignores \emph{higher level dependencies} that could be present in a tensor-structured output. 
For illustration, suppose the output is a matrix encoding the samples of $d_1$ continuous variables at $d_2$ different time steps,
 one could expect structural relations between the $d_1$ time series, for example linear dependencies between the rows 
 of the output matrix.

\paragraph{Low-rank regression for tensor responses.} 
To overcome the limitation described above we propose an  extension of the low-rank regression method 
 for tensor-structured responses by enforcing low multilinear rank of the regression tensor $\ten{W}$. 
Let $\{(\vec{x}\nsample{n}, \ten{Y}\nsample{n})\}_{n=1}^N 
\subset \R^{d_0}\times \R^{d_1\times d_2\times \cdots\times d_p}$ be  a  training sample of input/output data
drawn from the model $f(\vec{x}) = \ten{W} \ttv {1}\vec{x} + \ten{E}$ where $\ten{W}$ is assumed of low
multilinear rank. 
Considering the framework of empirical risk minimization, we want to find a low-rank regression tensor 
$\W$ minimizing the loss on the training data. To avoid numerical 
instabilities and to prevent overfitting we add a ridge regularization to the objective
function, leading to the following minimization problem
\begin{align}
\label{eq:holrr-general-loss}
\MoveEqLeft\min_{\W\in \R^{d_0\times  \cdots \times d_p}} \sum_{n=1}^N
\ell(\W\ttv{1}\vec{x}\nsample{n}, \ten{Y}\nsample{n}) + \gamma \norm{\ten{W}}^2_F \\
 &\text{ s.t. } \rank(\W) \leq (R_0,R_1,\cdots,R_p)\nonumber\enspace,
\end{align}
for some given integers $R_0,R_1,\cdots,R_p$ and where $\ell$ is a loss function. 
In this paper, we consider the squared error loss
between tensors defined by
$\ell(\ten{T},\hat{\ten{T}}) = \norm{\ten{T}-\hat{\ten{T}}}^2_F$.  
Using this loss we can rewrite problem~(\ref{eq:holrr-general-loss}) as
\begin{align}\label{eq:holrr-pbm}
\MoveEqLeft\min_{\W\in \R^{d_0\times d_1 \times \cdots \times d_p}} \norm{\W\ttm{1}\X - \Y}^2_F 
+ \gamma \norm{\W}^2_F \\
&\text{s.t. }\rank(\W) \leq  (R_0, R_1, \cdots, R_p)\enspace,\nonumber
\end{align}
where the input matrix $\X\in \R^{N\times d_0}$ and the output tensor
$\Y\in \R^{N\times d_1 \times \cdots \times d_p}$ are defined by
$\X_{n,:} = (\vec{x}\nsample{n})^\top$, $\Y_{n,:,\cdots,:} = \Y\nsample{n}$
for $n=1,\cdots, N$ ($\Y$ is the tensor obtained by stacking the output
tensors along the first mode).

\paragraph{Low-rank regression function.} Let $\W^*$ be a solution of problem~(\ref{eq:holrr-pbm}), it
follows from the multilinear rank constraint that $\W^* = \G \ttm{1}\U_0\ttm{2}\cdots\ttm{p+1}\U_p$
for some core tensor $\G \in \R^{R_0\times\cdots\times R_p}$ and orthogonal matrices 
$\U_i\in\R^{d_i\times R_i}$ for $0\leq i \leq p$. The regression function $f^* : \vec{x} \mapsto \W^*\ttv{1}\vec{x}$
can thus be written as 
$$f^*:\x \mapsto \G \ttm{1}\x^\top\U_0\ttm{2}\cdots\ttm{p+1}\U_p\enspace .$$

This implies several interesting properties. First, for any $\vec{x}\in \R^{d_0}$ we
have $f^*(\x) = \T_\x \ttm{1}\U_1\ttm{2}\cdots\ttm{p} \U_p$ with 
$\T_\x = \G \ttv{1} \U_0^\top\x$, which implies $\rank(f^*(\vec{x})) \leq (R_1,\cdots,R_p)$,
that is the image of $f^*$ is a set of tensors with low multilinear rank. Second, the relation
between $\x$ and $\Y=f^*(\x)$ is explained in a low dimensional subspace of size 
$R_0\times R_1\times\cdots\times R_p$. Indeed one can decompose the mapping $f^*$ into the 
following steps: (i) project $\x$ in $\R^{R_0} $ as $\bar{\x} = \U_0^\top\x$, (ii) perform
a low-dimensional mapping  $\bar{\Y} = \G\ttv{1}\bar{\x}$, (iii) project back into the
output space to get $\Y = \bar{\Y}\ttm{1}\U_1\ttm{2}\cdots\ttm{p}\U_p$.

\paragraph{Comparison with LRR.} First, it is obvious that the function learned by vectorizing
the outputs and performing LRR does not enforce any multilinear structure in the predictions.
That is, the tensor obtained by reshaping a vectorized prediction will not necessarily have a 
low multilinear rank. This is well illustrated in the experiment from Section~\ref{sec:xp_images}
where this fact is particularly striking when we try to learn an image from noisy measurements 
with a rank constraint set to $1$ (see Figure~\ref{fig:xp_images}, left).
Second, since the multilinear rank is defined with respect to the ranks of the matricizations 
there are some scenarios where the rank constraint in LRR after vectorizing the output will 
be (almost) equivalent to the multilinear rank constraint in~(\ref{eq:holrr-pbm}). 
This is clear if the regression tensor is of low mode-$1$ rank and of full mode-$n$ rank for
$n > 1$, but even when $R_0\ll R_i$ for $i\geq 1$ we can expect  LRR to be
able to capture most of the low-rank structure which is concentrated along the first mode.


\subsection{Higher-Order Low-Rank Regression}
\label{sec:holrr}
We now propose an efficient algorithm to tackle problem~(\ref{eq:holrr-pbm}).
Theoretical approximation guarantees are given in the next section.

\paragraph{Problem reformulation.}  We first show that the ridge regularization term in
~(\ref{eq:holrr-pbm}) can be incorporated in the data fitting term.
Let $\tX\in \R^{(N+d_0)\times d_0}$ and $\tY\in \R^{(N+d_0)\times d_1 \times \cdots \times d_p}$
be defined by $\tX^\top = \left(\X\ |\ \gamma\I\right)^\top$ and 
$\tenmat{\tilde{Y}}{1}^\top = \left(\tenmat{Y}{1}\ |\ \mats{0}\right)^\top$.
It is easy to check that the objective function in~(\ref{eq:holrr-pbm}) is equal to 
$\norm{\W\ttm{1}\tX - \tY}^2_F$. Minimization problem~(\ref{eq:holrr-pbm})
can then be rewritten as
\begin{align}\label{eq:holrr-pbm2}
\MoveEqLeft\hspace{-1.5cm}\min_{\G\in \R^{R_0\times R_1 \times \cdots \times R_p},
\atop \U_i \in \R^{d_i\times R_i} \text{ for } 0\leq i \leq p} 
\norm{\W\ttm{1}\tX - \tY}^2_F  \\
\quad\text{s.t. }&\W = \G\ttm{1}\U_0\ttm{2}\U_1\cdots\ttm{p+1}\U_p,\nonumber\\
\quad\ \ \ \ \ \ &\U_i^\top\U_i=\I \text{  for } 0\leq i \leq p\enspace.\nonumber
\end{align}

We now show that this minimization problem can be reduced to finding $p+1$
projection matrices onto subspaces of dimension $R_0,R_1,\cdots, R_p$.
We start by showing that the core tensor $\G$ solution  of~(\ref{eq:holrr-pbm2})
is determined by the factor matrices $\U_0,\cdots,\U_p$. 
\begin{theorem}\label{thm:core-tensor}
For given orthogonal matrices $\U_0,\cdots,\U_p$ the tensor $\G$ that
minimizes~(\ref{eq:holrr-pbm2}) is given by
$$\G = \tY \ttm{1}(\U_0^\top\tX^\top\tX\U_0)^{-1}\U_0^\top\tX^\top
\ttm{2}\U_1^\top\ttm{3}\cdots\ttm{p+1}\U_p^\top\enspace.$$
\end{theorem}
\begin{proof}
Using Eq.~\ref{eq:vecTucker} the objective function in~(\ref{eq:holrr-pbm2}) can be written as
$$\norm{(\U_p\kron \U_{p-1}\kron\cdots\kron\U_1\kron\tX\U_0)\vectorize{\G} - \vectorize{\tY}}^2_F\enspace.$$
Let $\M = \U_p\kron \U_{p-1}\kron\cdots\kron\U_1\kron\tX\U_0$. The solution  w.r.t. $\vectorize{\G} $ of this
classical linear least-squares problem is given by $(\M^\top\M)^{-1}\M^\top$. Using the 
mixed-product and inverse properties of the Kronecker product and the column-wise orthogonality of
$\U_1,\cdots,\U_p$ we obtain
$$ \vectorize{\G} = \left(\U_p\kron\cdots\kron\U_1\kron
(\U_0^\top\tX^\top\tX\U_0)^{-1}\U_0^\top\tX^\top\right)\vectorize{\tY}\enspace.\qedhere$$
\end{proof}

It follows from Theorem~\ref{thm:core-tensor} that problem~(\ref{eq:holrr-pbm}) 
can be written as 
\begin{align}\label{eq:holrr-pbm3}
\min_{\U_i \in \R^{d_i\times R_i},\atop 0\leq i \leq p} &
\norm{\tY\ttm{1}\mats{\Pi}_0\ttm{2}\cdots\ttm{p+1}\mats{\Pi}_p - \tY}^2_F  \\
\text{\ \ \ \ s.t. }&\U_i^\top\U_i=\I\text{ for } 0\leq i \leq p\nonumber,\\
&\mats{\Pi}_0 = \tX\U_0\left(\U_0\tX^\top\tX\U_0^\top\right)^{-1}\U_0^\top\tX^T,\nonumber\\
&\mats{\Pi}_i = \U_i\U_i^\top \text{ for }1\leq i \leq p\enspace,\nonumber
\end{align}
where we used $ \W\ttm{1}\tX= \tY\ttm{1}\mats{\Pi}_0\ttm{2}\cdots\ttm{p+1}\mats{\Pi}_p$.
Note that $\mats{\Pi}_0$ is the orthogonal projection onto the space spanned by the
columns of $\tX\U_0$ and $\mats{\Pi}_i$ is the orthogonal projection onto the column
space of $\U_i$ for $i\geq 1$.
Hence solving problem~(\ref{eq:holrr-pbm}) is equivalent to finding $p+1$ low-dimensional 
subspaces $U_0,\cdots,U_p$ such that projecting  $\tY$ onto
the spaces $\tX U_0, U_1,\cdots, U_p$ along the corresponding modes is  close to $\tY$.

\paragraph{HOLRR algorithm.}
Since solving problem~(\ref{eq:holrr-pbm3}) for the $p+1$ projections simultaneously is a difficult
non-convex optimization problem we propose to solve it independently for each projection. This 
approach has the benefits of both being computationally efficient and providing good theoretical
approximation guarantees (see Theorem~\ref{thm:approx}). The following proposition gives the
analytic solutions of~(\ref{eq:holrr-pbm3}) when  each projection is considered independently.

\begin{proposition}\label{prop:projection-sol}
For $0\leq i \leq p$, using the definition of $\mats{\Pi}_i$ in~(\ref{eq:holrr-pbm3}), the optimal solution of 
$$\min_{\U_i \in \R^{d_i\times R_i}  }\norm{\tY\ttm{i+1}\mats{\Pi}_i - \tY}^2_F \text{ s.t. } \U_i^\top\U_i = \I$$
is given by the eigenvectors of 
$$
\begin{cases}
(\tX^\top\tX) ^{-1}\tX^\top\tenmat{\tilde{Y}}{1}\tenmat{\tilde{Y}}{1}^\top\tX &\text{ if } i=0\\
\tenmat{\tilde{Y}}{i}\tenmat{\tilde{Y}}{i}^\top &\text{ otherwise}
\end{cases}
$$
that corresponds to the $R_i$ largest eigenvalues.
\end{proposition}
\begin{proof}
For any $0\leq i \leq p$, since $\mats{\Pi}_i$ is a projection we have
$\langle \tY\ttm{1} \mats{\Pi}_i, \tY\rangle = \langle \mats{\Pi}_i\tenmat{\tilde{Y}}{i}, \tenmat{\tilde{Y}}{i}\rangle
= \norm{\mats{\Pi}_i\tenmat{\tilde{Y}}{i}}^2_F$, thus minimizing $\norm{ \tilde{\Y}\ttm{i}\mats{\Pi}_i - \tilde{\Y}}^2_F$
is equivalent to minimizing 
$  \norm{\mats{\Pi}_i\tenmat{\tilde{Y}}{i}}^2_F - 2\langle \mats{\Pi}_i\tenmat{\tilde{Y}}{i}, \tenmat{\tilde{Y}}{i}\rangle
= - \norm{\mats{\Pi}_i\tenmat{\tilde{Y}}{i}}^2_F$.
For $i \geq 1$, we have $\norm{\mats{\Pi}_i\tenmat{\tilde{Y}}{i}}^2_F = \Tr(\U_i^\top \tenmat{\tilde{Y}}{i}\tenmat{\tilde{Y}}{i}^\top \U_i)$
which is maximized by letting the columns of $\U_i$ be the top $R_i$ eigenvectors of
the matrix  $\tenmat{\tilde{Y}}{i}\tenmat{\tilde{Y}}{i}^\top$. 
%
For $i=0$ we have  
$$
\norm{\mats{\Pi}_0\tenmat{\tilde{Y}}{i}}^2_F = \Tr(\mats{{\Pi}}_0\tenmat{\tilde{Y}}{1}\tenmat{\tilde{Y}}{1}^\top
{\mats{\Pi}}_0^\top)
 =
\Tr\left((\U_0^\top \mat{A} \U_0)^{-1} 
\U_0^\top \mat{B} \U_0\right)
$$
with $\mat{A} = \mat{\tX^\top\tX} $ and 
$\mat{B} =\tX^\top\tenmat{\tilde{Y}}{1}\tenmat{\tilde{Y}}{1}^\top\tX$,
which is maximized by the top $R_0$  eigenvectors of $\mat{A}^{-1}\mat{B}$. 
\end{proof}

The results from Theorem~\ref{thm:core-tensor} and Proposition~\ref{prop:projection-sol} can be
rewritten using the original input matrix $\X$ and output tensor $\Y$.
Since $\tX^\top\tX = \X^\top\X+\gamma\I$ and $\tilde{\ten{Y}}\ttm{1}\tX^\top=\ten{Y}\ttm{1}\X^\top$,
we can rewrite the solution of Theorem~\ref{thm:core-tensor} as 
$$\G = \Y \ttm{1}(\U_0^\top(\X^\top\X+\gamma\I)\U_0)^{-1}\U_0^\top\X^T
\ttm{2}\U_1^\top\ttm{3}\cdots\ttm{p+1}\U_p^\top\enspace.$$ Similarly for
Proposition~\ref{prop:projection-sol} we have
$$(\tX^\top\tX) ^{-1}\tX^\top\tenmat{\tilde{Y}}{1}\tenmat{\tilde{Y}}{1}^\top\tX =
(\X^\top\X + \gamma\I)^{-1}\X^\top\tenmat{Y}{1}\tenmat{Y}{1}^\top\X$$ and one
can check that $\tenmat{\tilde{Y}}{i}\tenmat{\tilde{Y}}{i}^\top = \tenmat{Y}{i}\tenmat{Y}{i}^\top$
for any $i \geq 1$.

The overall Higher-Order Low-Rank Regression procedure (HOLRR)  is summarized in Algorithm~\ref{alg:holrr}.
Note that the Tucker decomposition of the solution returned by HOLRR could be a good initialization point
for an Alternative Least Square method. However, studying the theoretical and experimental properties of
this approach is beyond the scope of this paper and is left for future work.

\begin{algorithm}
   \caption{\texttt{HOLRR}}
   \label{alg:holrr}
\begin{algorithmic}[1]
   \REQUIRE $\mat{X}\in \R^{N \times d_0}$, 
   $\Y \in \R^{N\times d_1\times\cdots\times d_p}$,
   rank $(R_0,R_1,\cdots,R_p)$ and
   regularization parameter $\gamma$.
   \STATE$\U_0\ \leftarrow$ top $R_0$  eigenvectors of\newline
${\hspace{1.5cm} \scriptsize (\X^\top\X + \gamma\I)^{-1}\X^\top\tenmat{Y}{1}\tenmat{Y}{1}^\top\X}$\label{alg:holrr-U0} 
   \FOR{$i=1$ {\bfseries to} $p$}
   \STATE$\U_i \leftarrow$ top $R_i$ eigenvectors of $\tenmat{Y}{i}\tenmat{Y}{i}^\top$\label{alg:holrr-Ui} 
   \ENDFOR
   \STATE $\mat{M} = \left(\U_0^\top(\X^\top\X+\gamma\I)\U_0\right)^{-1}\U_0^\top\X^\top$
	\STATE $\G \leftarrow \Y \ttm{1}\mat{M}\ttm{2}\U_1^\top\ttm{3}\cdots\ttm{p+1}\U_p^\top$
	\STATE  {\bfseries return}  $\G\ttm{1}\U_0\ttm{2}\cdots\ttm{p+1}\U_p$
\end{algorithmic}
\end{algorithm}


\subsection{Theoretical Analysis}

\paragraph{Complexity analysis.} We compare the computational complexity
of the LRR and  HOLRR algorithms.
For both algorithms the computational cost for matrix multiplications is asymptotically the same
(dominated by the products $\X^\top\X$ and $\tenmat{Y}{1}\tenmat{Y}{1}^\top$). 
We thus focus our analysis on the computational cost of
matrix inversions and truncated singular value decompositions. For the two methods
the inversion of the matrix $\X^\top\X +\gamma\I$ can be done in $\bigo{(d_0)^3}$. The LRR
method needs to compute the $R$ dominant eigenvectors of a matrix of
the same size as $\tenmat{Y}{1}^\top \tenmat{Y}{1}$ (see Section~\ref{sec:LRR})
which can be done in $\bigo{R(d_1d_2\cdots d_p)^2}$. The HOLRR algorithm needs
to compute the $R_i$ dominant eigenvectors of the matrix $\tenmat{Y}{i} \tenmat{Y}{i}^\top$
for $i\in [p]$ and to invert the extra matrix $\U_0^\top(\X^\top\X+\gamma\I)\U_0$ of
size $R_0\times R_0$ leading to a complexity of $\bigo{(R_0)^3 + \max_i\{R_id_i^2\}}$. 
We can conclude that 
HOLRR is far more efficient than LRR when the output dimensions $d_1,\cdots,d_p$ are large.

\paragraph{Approximation guarantees.}
The following theorem shows that the HOLRR algorithm proposed in the previous section
is a $(p+1)$-approximation algorithm for problem~(\ref{eq:holrr-pbm}).
\begin{theorem}
\label{thm:approx}
Let $\W^*$ be a solution of problem~(\ref{eq:holrr-pbm}) and
let $\W$ be the regression tensor  returned by Algorithm~\ref{alg:holrr}.
If $\mathcal{L}:\R^{d_0\times \cdots \times d_p } \to \R$ denotes the objective 
function of~(\ref{eq:holrr-pbm}) w.r.t. $\W$ then
$\mathcal{L}(\W) \leq  (p+1) \mathcal{L}(\W^*) .$
\end{theorem}
\begin{proof}
Let $\U_0,\cdots,\U_p$ be the matrices defined in Algorithm~\ref{alg:holrr} and
let $\mats{\Pi}_0,\cdots,\mats{\Pi}_p$ be the projection matrices defined in
problem~(\ref{eq:holrr-pbm3}).
We have

\begin{align*}
\mathcal{L}(\W) &= \norm{\W\ttm{1}\tilde{\X} - \tilde{\Y}}^2_F\\
&= \norm{\tY\ttm{1}\mats{\Pi}_0\ttm{2}\cdots\ttm{p+1}\mats{\Pi}_p - \tY}^2_F \\
&\leq \norm{\tY\ttm{1}\mats{\Pi}_0\ttm{2}\cdots\ttm{p+1}\mats{\Pi}_p 
                     - \tY\ttm{1}\mats{\Pi}_0\ttm{2}\cdots\ttm{p}\mats{\Pi}_{p-1}}^2_F \\
&+ \norm{\tY\ttm{1}\mats{\Pi}_0\ttm{2}\cdots\ttm{p}\mats{\Pi}_{p-1}
               - \tY\ttm{1}\mats{\Pi}_0\ttm{2}\cdots\ttm{p-1}\mats{\Pi}_{p-2}}^2_F \\
 &+\cdots 
 + \norm{\tY\ttm{1}\mats{\Pi}_0 - \tY}^2_F\\
&\leq \norm{\tY\ttm{p+1}\mats{\Pi}_p  - \tY}^2_F + \norm{\tY\ttm{p}\mats{\Pi}_{p-1}-\tY}^2_F
+ \cdots +   \norm{\tY\ttm{1}\mats{\Pi}_0 - \tY}^2_F\enspace,
\end{align*}
where we used the fact that $\norm{\mats{\Pi}\M}_F \leq \norm{\M}_F$ for any projection matrix $\mats{\Pi}$.
Furthermore, it follows from Theorem~\ref{thm:core-tensor} that 
$\W^* = \Y \ttm{1}\mats{\Pi}_0^* \ttm{2} \cdots \ttm{p+1}\mats{\Pi}_p^*$
for some projection matrices $\mats{\Pi}_i^*$ satisfying the constraints from problem~(\ref{eq:holrr-pbm3}).
By Proposition~\ref{prop:projection-sol} each summand in the last inequality is minimal w.r.t. $\mats{\Pi}_i$, hence
\begin{align*}
\norm{\tY\ttm{i}\mats{\Pi}_i - \tY}^2_F &\leq \norm{\tY\ttm{i}\mats{\Pi}^*_i - \tY}^2_F\\
&\leq  \norm{\tY\ttm{1}\mats{\Pi}^*_0\ttm{2}\cdots \ttm{p+1}\mats{\Pi}^*_p - \tY}^2_F \\
 &= \norm{\W^*\ttm{1}\tX- \tY}^2_F \\
 &= \mathcal{L}(\W^*)
\end{align*}
which concludes the proof.
\end{proof}

One direct consequence of this theorem is that if there exists a $\W^*$ such
that $\W^*\ttm{1}\X = \Y$ then Algorithm~\ref{alg:holrr} (using $\gamma=0$)
will return a tensor $\W$ satisfying $\W\ttm{1}\X = \Y$.


\section{HOLRR Kernel Extension}
\label{sec:kernel}
In this section we provide a kernelized version of the HOLRR algorithm. We proceed by analyzing
how the algorithm would be instantiated in a feature space and we show that all the 
steps involved can be performed using the Gram matrix of the input data without
having to explicitly compute the feature map. 

Let $\phi: \R^{d_0} \to \R^L$ be a feature map and let $\mats{\Phi} \in \R^{n\times L}$ be the
matrix with lines $\phi(\vec{x}\nsample{n})^\top$ for $n\in[N]$. The higher-order low-rank 
regression problem in the feature space boils down to the minimization problem
\begin{align}\label{eq:kholrr-pbm}
\MoveEqLeft\min_{\W\in \R^{L\times d_1 \times \cdots \times d_p}} \norm{\W\ttm{1}\mats{\Phi} - \Y}^2_F 
+ \gamma \norm{\W}^2_F \\
&\text{s.t. }\rank(\W) \leq  (R_0, R_1, \cdots, R_p)\enspace.\nonumber
\end{align}

Following the HOLRR algorithm proposed in Section~\ref{sec:holrr} one needs to compute
the top $R_0$ eigenvectors of the $L\times L$ matrix
$( \mats{\Phi}^\top\mats{\Phi}  + \gamma\I)^{-1}\mats{\Phi}^\top\tenmat{Y}{1}\tenmat{Y}{1}^\top\mats{\Phi} $.
The following proposition shows that this can be done using the Gram matrix $\K = \mats{\Phi} \mats{\Phi} ^\top$
without having to explicitly  compute the feature map $\phi$.

\begin{proposition}
If $\vecs{\alpha}\in\R^N$ is an eigenvector with eigenvalue $\lambda$ of the matrix 
\begin{equation}
\label{eq:gram-eigvec}
(\K+\gamma\I)^{-1}\tenmat{Y}{1}\tenmat{Y}{1}^\top \K\enspace,
\end{equation}
then $\vec{v} = \mats{\Phi}^\top \vecs{\alpha} \in\R^L$ is an eigenvector with eigenvalue $\lambda$ of
the matrix
$( \mats{\Phi}^\top\mats{\Phi}  + \gamma\I)^{-1}\mats{\Phi}^\top\tenmat{Y}{1}\tenmat{Y}{1}^\top\mats{\Phi} .$
\end{proposition}
\begin{proof}
Let  $\vecs{\alpha}\in\R^N$ be the eigenvector from the hypothesis. We have
\begin{align*}
\lambda\vec{v}
&=
\mats{\Phi}^\top (\lambda \vecs{\alpha}) 
=
\mats{\Phi}^\top\left( (\K+\gamma\I)^{-1}\tenmat{Y}{1}\tenmat{Y}{1}^\top \K \right)\vecs{\alpha}\\
&=
\mats{\Phi}^\top( \mats{\Phi} \mats{\Phi}^\top + \gamma\I)^{-1}
\tenmat{Y}{1}\tenmat{Y}{1}^\top\mats{\Phi} \mats{\Phi}^\top \vecs{\alpha}\\
&=
\left(( \mats{\Phi}^\top\mats{\Phi}  + \gamma\I)^{-1}
\mats{\Phi}^\top\tenmat{Y}{1}\tenmat{Y}{1}^\top\mats{\Phi} \right)\vec{v} \enspace.\qedhere
\end{align*}
\end{proof}

Let $\mat{A}$ be the top $R_0$ eigenvectors of the matrix~(\ref{eq:gram-eigvec}).
When working with the feature map $\phi$, it follows from the previous proposition that
 line~1 in Algorithm~\ref{alg:holrr} is  equivalent
to choosing $\U_0 = \mats{\Phi}^\top \mat{A}\in \R^{L\times R_0}$, while the updates
in line~3 stay the same. The regression tensor $\W\in\R^{L\times d_1\times\cdots\times d_p}$
 returned by this algorithm  is then equal to
\begin{align*}
\W = \Y\ttm{1}  \mat{P} \ttm{2} \U_1\U_1^\top \ttm{2} \cdots \ttm{p+1} \U_p\U_p^\top\enspace,
\end{align*}
where 
\begin{align*}
\mat{P} 
&= 
\mats{\Phi}^\top\mat{A}\left(\mat{A}^\top\mats{\Phi}
(\mats{\Phi}^\top\mats{\Phi}+\gamma\I) \mats{\Phi}^\top\mat{A}\right)^{-1}\mat{A}^\top\mats{\Phi}\mats{\Phi}^\top\\
&=
\mats{\Phi}^\top\mat{A}\left(\mat{A}^\top\mats{\Phi}\mats{\Phi}^\top
(\mats{\Phi} \mats{\Phi}^\top+\gamma\I)\mat{A}\right)^{-1}\mat{A}^\top\mats{\Phi}\mats{\Phi}^\top\\
&=
\mats{\Phi}^\top\mat{A}\left(\mat{A}^\top\K
(\K+\gamma\I)\mat{A}\right)^{-1}\mat{A}^\top\K\enspace.
\end{align*}

Suppose now that the feature map $\phi$ is induced by a kernel $k:\R^{d_0}\times\R^{d_0}\to \R$.
The prediction for an input vector $\vec{x}$ is then given by 
$\W\ttv{1}\vec{x} = \ten{C}\ttv{1}\vec{k}_{\vec{x}}$ where
the $n$th component of $\vec{k}_{\vec{x}}\in \R^N$ is $k(\vec{x}_n,\vec{x})$ and
the tensor $\ten{C}\in\R^{N\times  d_1\times \cdots \times d_p}$ is defined by
$$\ten{C} = \G \ttm{1}\mat{A}\ttm{2}\U_1\ttm{2}\cdots\ttm{p+1}\U_p^\top\enspace,$$ 
with
$\G = \Y\ttm{1} \left(\mat{A}^\top\K(\K+\gamma\I)\mat{A}\right)^{-1}\mat{A}^\top\K
\ttm{2}\U_2^\top\ttm{3}\cdots\ttm{p+1}\U_p$. 

Now, let $\mathcal{H}$ be the reproducing kernel Hilbert space associated with the kernel $k$.
The overall procedure for kernelized HOLRR is summarized in Algorithm~\ref{alg:kholrr}. This
algorithm returns the tensor $\ten{C} \in \R^{N\times d_1 \times \cdots \times d_p}$ defining 
the regression function 
%
%
$$f: \vec{x} \mapsto \ten{C}\ttv{1}\vec{k}_{\vec{x}} = \sum_{n=1}^N k(\vec{x},\vec{x}^{(n)}) \ten{C}^{(n)}\enspace,$$
%
%
where $\ten{C}^{(n)} = \ten{C}_{n:\cdots:}\in \R^{d_1\times\cdots\times d_p}$.

\begin{algorithm}
   \caption{\texttt{Kernelized HOLRR}}
   \label{alg:kholrr}
\begin{algorithmic}[1]
   \REQUIRE Gram matrix $\mat{K}\in \R^{N \times N}$, 
   $\Y \in \R^{N\times d_1\times\cdots\times d_p}$,
   rank $(R_0,R_1,\cdots,R_p)$ and
   regularization parameter $\gamma$.
   \STATE\label{alg:holrr-U0} $\mat{A}\ \leftarrow$ top $R_0$  eigenvectors of\newline
${\hspace{1.5cm} \scriptsize (\K+\gamma\I)^{-1}\tenmat{Y}{1}\tenmat{Y}{1}^\top \K}$
   \FOR{$i=1$ {\bfseries to} $p$}
   \STATE\label{alg:holrr-U0} $\U_i \leftarrow$ top $R_i$ eigenvectors of $\tenmat{Y}{i}\tenmat{Y}{i}^\top$
   \ENDFOR
   \STATE $\mat{M}\leftarrow \left(\mat{A}^\top\K(\K+\gamma\I)\mat{A}\right)^{-1}\mat{A}^\top\K$
	\STATE $\G \leftarrow \Y \ttm{1}\mat{M} \ttm{2}\U_1^\top\ttm{3}\cdots\ttm{p+1}\U_p^\top$
	\STATE  {\bfseries return}  $\ten{C} = \G\ttm{1}\mat{A}\ttm{2}\U_1\ttm{3}\cdots\ttm{p+1}\U_p$
\end{algorithmic}
\end{algorithm}

 \section{Experiments}
\label{sec:xp}
We present and analyze the experimental results obtained on
a regression task on synthetic data, an
image reconstruction task and a meteorological forecasting task on real data%
\footnote{The code and data used in the experiments will be made available by the authors.}.
We compare the predictive accuracy of  HOLRR with the following methods:\newline
- RLS: Regularized least squares.\newline
- LRR: Low-rank regression (see Section~\ref{sec:LRR}).\newline
- ADMM: a multilinear approach based on tensor trace norm regularization introduced in \citep{gandy2011tensor}
 and \citep{romera2013}.\newline
 - MLMT-NC: a nonconvex approach proposed in \citep{romera2013}  in the context of multilinear multitask learning.
 
 For experiments with kernel algorithms we use the readily available kernelized RLS
  and the LRR kernel extension  proposed 
 in~\citep{Mukherjee2011} (note that ADMM and MLMT-NC only consider a linear
 dependency between inputs and outputs).
 
 Although  MLMLT-NC is perhaps the closest algorithm to ours, we applied it only to simulated data. 
 This is because MLMLT-NC is computationally  very expensive and becomes intractable for large data sets.
  To give an idea on the running times,
 averages for some of the experiments are reported in Table~\ref{tab:runtime}.
 
 \begin{table}
 \begin{center}
 
 \caption{Average running time for different size of the regression tensor $\W$. Each line
 comes from one of the experiments. The times reported
 are the average learning times in seconds for a training set of size $100$.}
 \label{tab:runtime}
  \begin{scriptsize}
  
 \vspace{0.1cm}
 
 \begin{sc}
 \setlength{\tabcolsep}{3pt}
 \begin{tabular}{c|ccccc}
 \hline
size of $\W$ & RLS & LRR & ADMM & MLMT-NC & HOLRR\\

\hline
\rule{0pt}{2.6ex}

${10\times 10\times 10\times 10}$ &$0.012$ & $0.45$ & $12.92$ & $945.79 $ & $0.04$ \\
${3\times 70\times 70}$ & $0.04$ & $10.46$ & $-$&$-$&$0.08$\\ 
${160\times 5\times 16 \times 5}$  & $2.07$ & $2.24$ & $40.23$ & $-$ & $1.67$ \\
\hline
 \end{tabular}
 \end{sc}
 \end{scriptsize}
 
 \vspace{-0.5cm}

 \end{center}
 \end{table}
 
\subsection{Synthetic Data}
We generate both linear and nonlinear data.
The linear data is drawn from the model  $\ten{Y} = \ten{W} \ttv{1} \vec{x} + \ten{E}$
where $\ten{W}\in \R^{10\times 10\times 10 \times 10}$ is a tensor of multilinear 
rank $(6,4,4,8)$ drawn at random, $\x\in\R^{10}$ is drawn from $\mathcal{N}(0,\I)$ and each component
of the error tensor $\ten{E}$ is drawn from $\mathcal{N}(0,0.1)$. The nonlinear data
is drawn from the model $\ten{Y} = \ten{W} \ttv{1} (\vec{x}\kron\x) + \ten{E}$ where
$\ten{W}\in \R^{25\times 10\times 10 \times 10}$ is a tensor of  
rank $(5,6,4,2)$ drawn at random and $\x\in\R^5$ and $\ten{E}$ are generated as above.
The hyper-parameters for all algorithms are selected using $3$-fold cross validation
on the training data. 

\begin{figure}

\begin{center}
\includegraphics[scale=0.45]{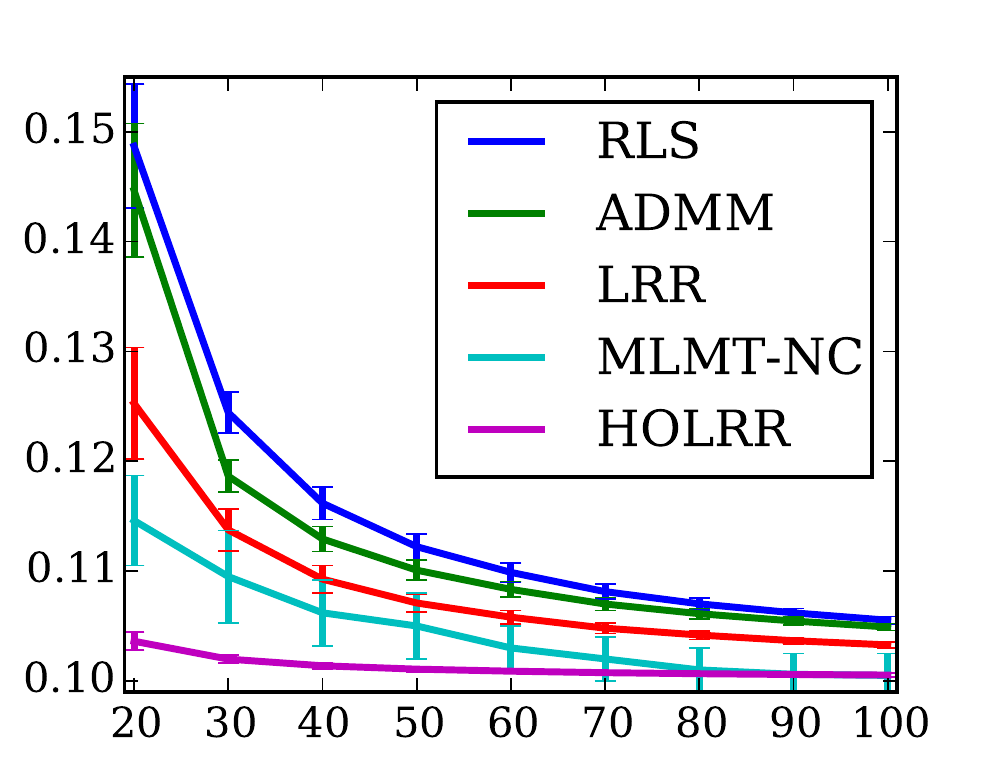}%
\includegraphics[scale=0.45,trim=1cm 0cm 0cm 0cm]{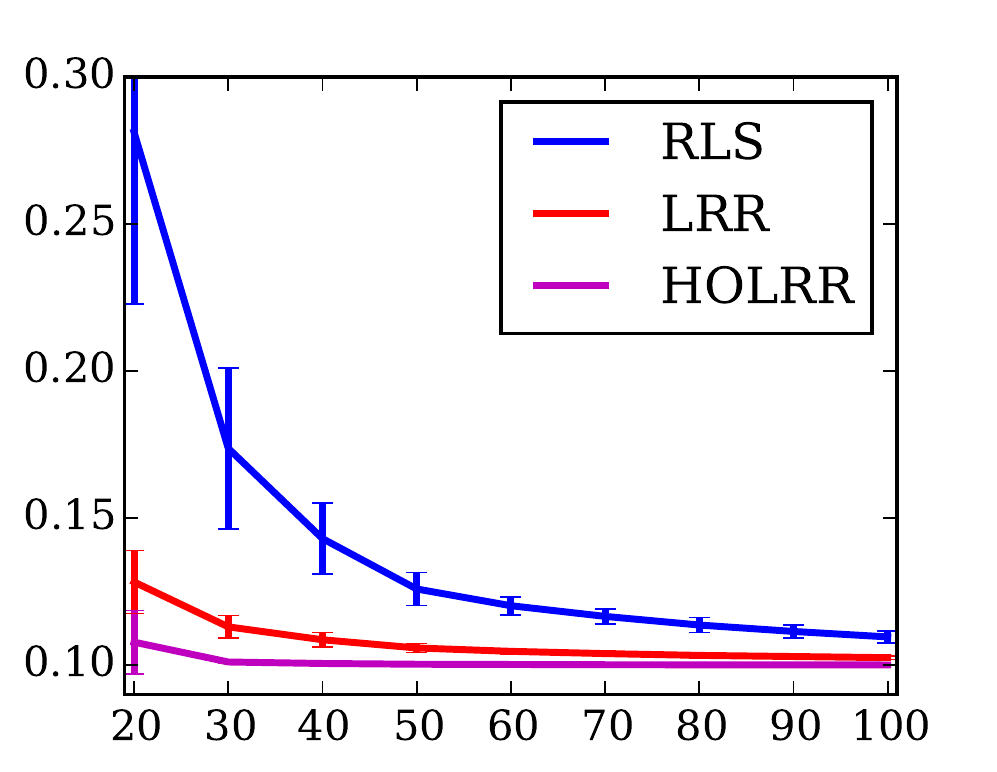}
\caption{Average RMSE as a function of the training set size: (left) linear data, (right) nonlinear data.}
\label{fig:xp_synth}
\end{center}

\vspace{-0.7cm}

\end{figure}

These experiments have been carried out for different sizes of the training data set, 20 trials have been executed for each size.  
The average RMSEs on a test set of size 100 for the 20 trials are reported in Figure~\ref{fig:xp_synth}. We see
that the HOLRR algorithm clearly outperforms the other methods on the linear data. MLMT-NC is the method
obtaining the better accuracy after ours, it is however much more computationally expensive%
~(see Table~\ref{tab:runtime}). On the nonlinear data the LRR method achieves good performances
but HOLRR is still significantly more accurate for small data set sizes.

\subsection{Image Reconstruction from Noisy Measurements}
\label{sec:xp_images}
To give an illustrative intuition on the differences between matrix and multilinear rank regularization
 we generate data  from the model $\ten{Y} = \ten{W} \ttv{1} \vec{x} + \ten{E}$ where
the tensor $\W$ is  a color image of size $m\times n$ encoded with three color channels
 RGB. We consider two different tasks depending on the input dimension: 
 (i) $\W\in\R^{3\times m \times n}$, $\x\in \R^3$ and 
 (ii) $\W\in\R^{ n \times m \times 3}$, $\x\in \R^n$.
 In both tasks the components of both $\x$  and $\ten{E}$ are drawn from $\mathcal{N}(0,1)$ and the 
 regression tensor $\W$ is learned from a training set of size $200$. 

\begin{figure*}[th]
\vspace{-1cm}
    \hspace{-3cm}\begin{minipage}{.5\textwidth}
        \centering
        \includegraphics[height=3.5cm,,trim=0cm 0cm 0cm 0cm]{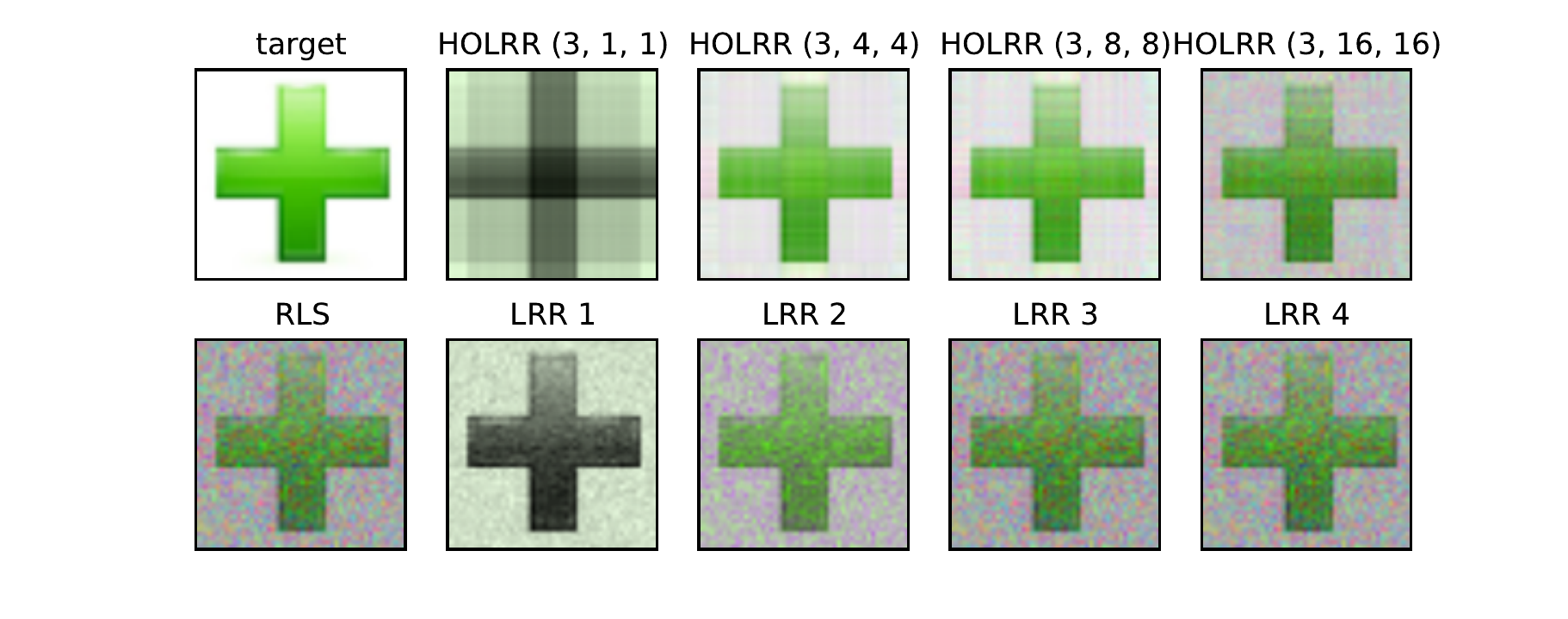}
		\includegraphics[height=4.5cm,,trim=0cm 0.5cm 0cm 0.5cm]{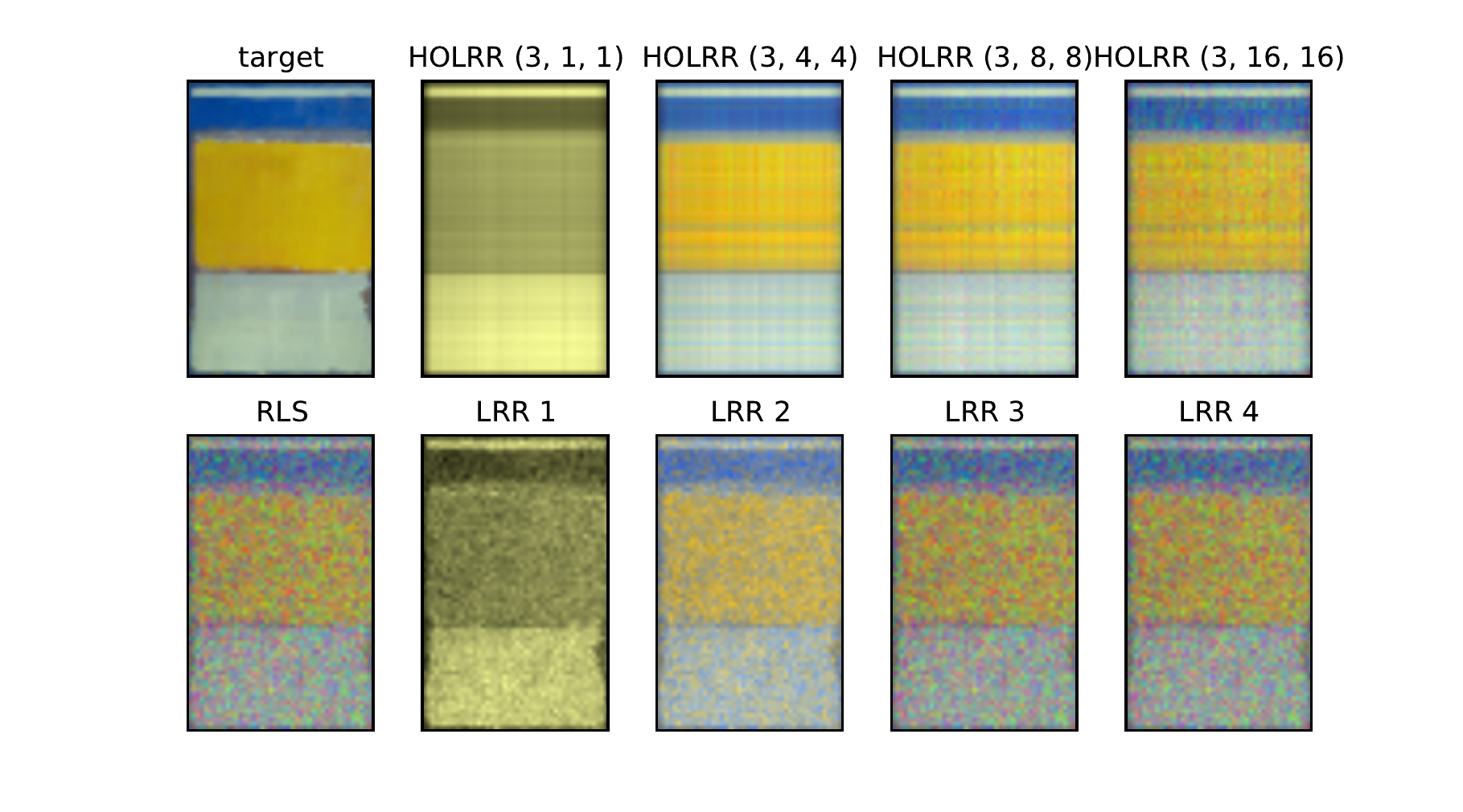}
		\includegraphics[height=3.5cm,,trim=0cm 0cm 0cm 0cm]{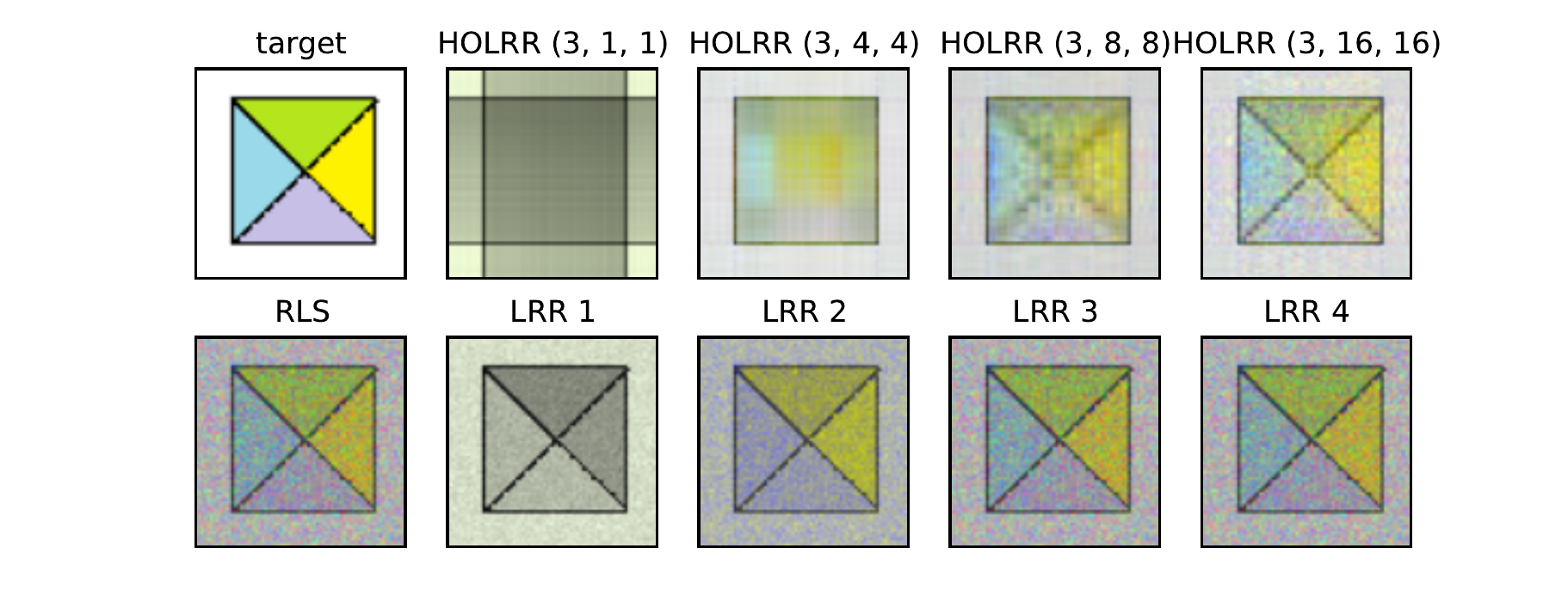}
    \end{minipage}\hspace{2.5cm}%
\begin{minipage}{.5\textwidth}
        \centering
        \includegraphics[height=3.5cm,trim=0cm 0cm 0cm 0cm]{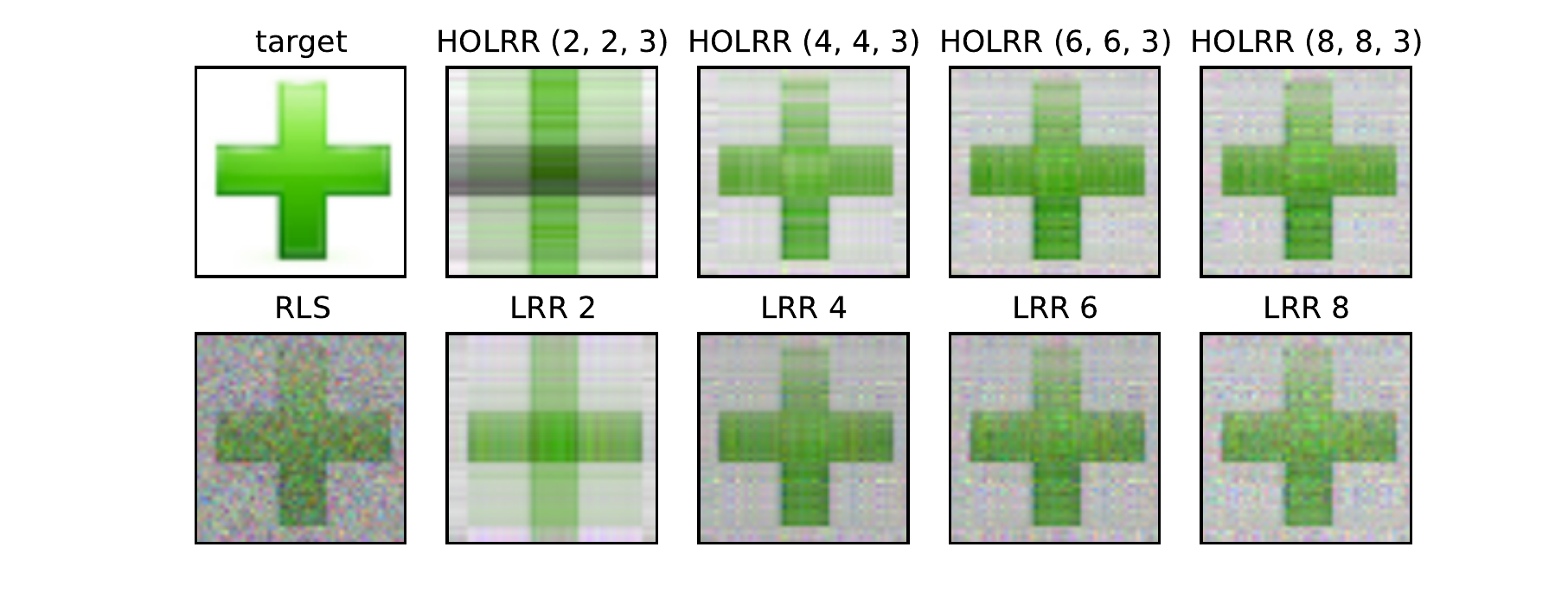}
		\includegraphics[height=4.5cm,,trim=0cm 0.5cm 0cm 0.5cm]{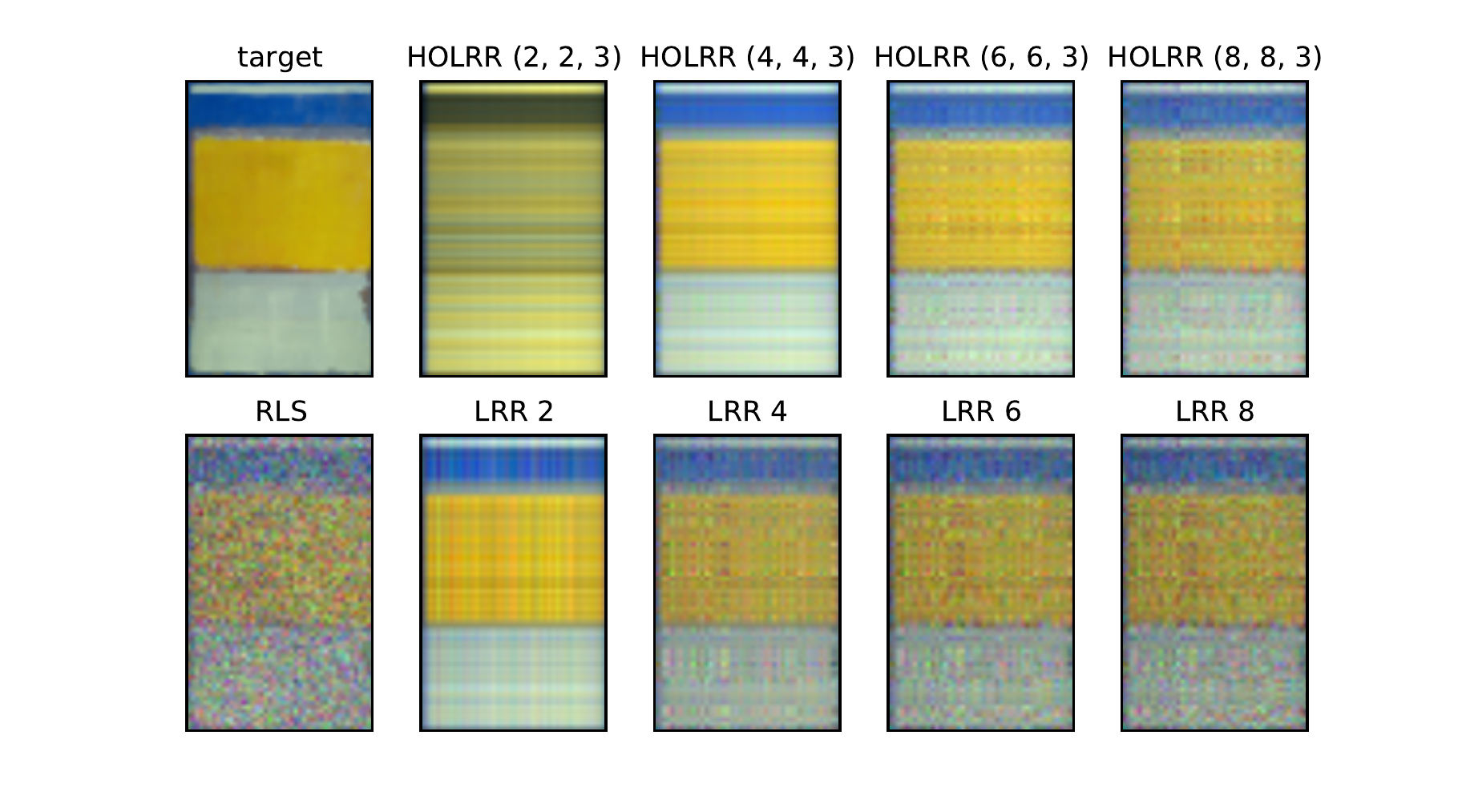}
		\includegraphics[height=3.5cm,,trim=0cm 0cm 0cm 0cm]{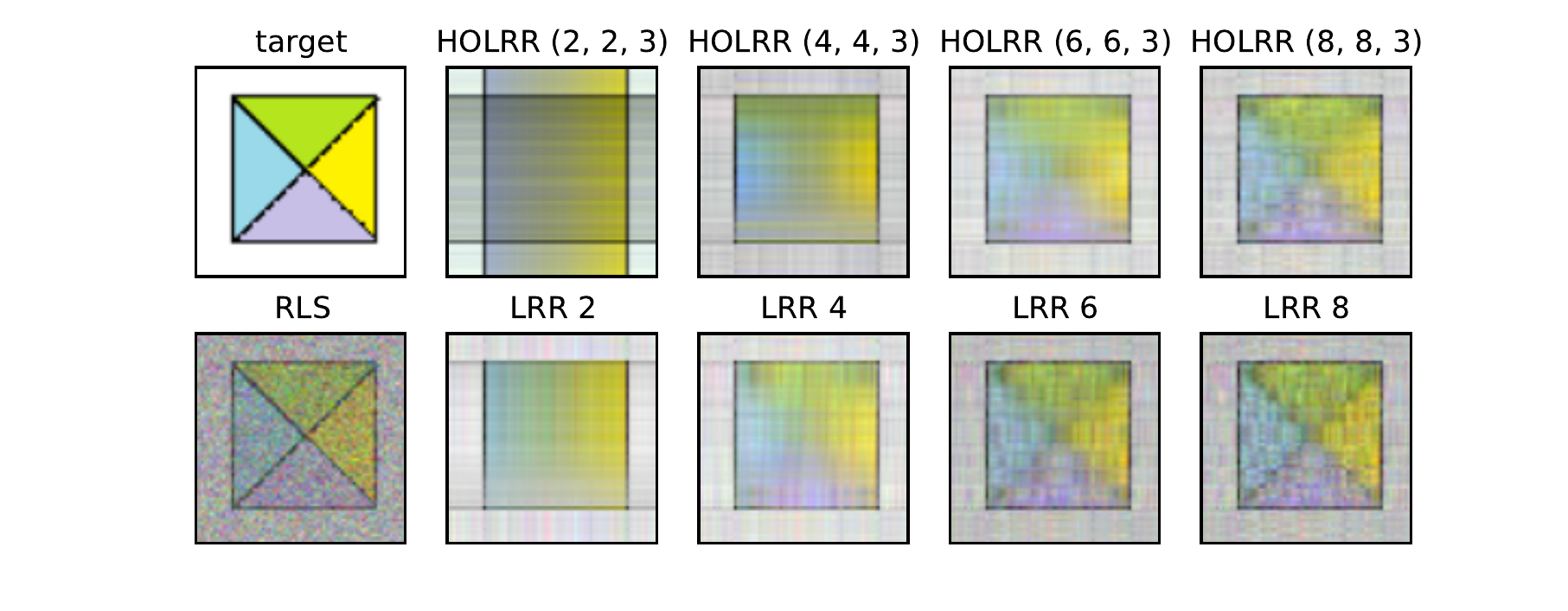}
    \end{minipage}
    \caption{Image reconstruction from noisy measurements: $\Y = \W\ttv{1}\vec{x} + \ten{E}$ where $\W$ is a color image (RGB). 
    (left) Task (i): input dimension is the number of channels. (right) Task (ii): input dimension is the height
    of the image. Each image is labeled with the name of the algorithm followed by the value used for the rank constraint.}
    \label{fig:xp_images}
\end{figure*}

 This experiment allows us to visualize the tensors returned by the RLS,  LRR and HOLRR algorithms.
 The results are shown in Figure~\ref{fig:xp_images} for three images: a green cross (of size
$50\times 50$), a thumbnail of a Rothko painting ($44\times 70$) and a square made of
triangles ($70\times 70$), note that the first two images have a low rank structure which
is not the case for the third one. 

We first see that HOLRR clearly outperforms LRR on the task where the input dimension is small (task (i)).
This is to be  expected since the rank of the matrix $\tenmat{W}{1}$ is at most $3$ and
LRR is unable to enforce a low-rank structure on the output modes of $\W$. When the rank constraint is
set to $1$ for LRR and $(3,1,1)$ for HOLRR we clearly see that unlike HOLRR 
the LRR approach does not enforce any low-rank
structure on the regression tensor along the output modes.
On task (ii) the difference is more subtle, but we can see that setting a rank constraint
of $2$ for the LRR algorithm prevents the model from capturing the white border around the green
cross and creates the vertical lines artifact in the Rothko painting. For higher values of the rank the
model starts to learn the noise. The tensor returned by HOLRR with rank $(4,4,3)$ does not exhibit these
behaviors and gives better results
on these two images. On the square image which does not have a low-rank structure both 
algorithms do not perform very well.
Overall, we see that capturing the mutlilinear low-rank structure of the output data allows HOLRR to
separate the noise from the true signal better than RLS and LRR.

\subsection{Real Data}
We compare our algorithm with other methods on the task of meteo forecasting. 
We collected data from the meteorological office of the UK
\footnote{\tiny{\url{http://www.metoffice.gov.uk/public/weather/climate-historic/}}}: 
monthly average measurements of 5 variables in 16 stations across the UK from 1960 to 2000.
 The forecasting task consists in predicting the values of the 5 variables in  the 16 stations from their values in the preceding months.  
 We use the values of all the variables from the past 2 months as covariates and consider
 the tasks of predicting the values of all variables for the next $k$ months,
 the output tensors are thus in $\R^{k\times 16 \times 5}$.
 
 We randomly split the available data into a training set of size $N$, a test set of size $50$ and
 a validation set of size $20$, for all methods hyper-parameters are chosen w.r.t. to their
 performance on the validation set. The average test RMSE over $10$ runs of this experiment 
 for different values of $N$ and $k$ are reported in Table~\ref{tab:xp_meteo}. We see that the
 HOLRR has overall a better predictive accuracy than the other methods, especially when the
 tensor structure of the output data gets richer ($k=3$ and $k=5$). On this dataset, nonlinear
 methods only improve the results by a small margin with the RBF kernel, and the polynomial
 kernel performs poorly.

 \begin{table}
 \begin{center}
 \caption{Forecasting experiment: the task is to predict the values for all variables for the next
 $k$ months using a training set of size $N$.}
 \label{tab:xp_meteo}
 \begin{scriptsize}
 
 \vspace{0.1cm}
 
 \begin{sc}
\begin{tabular}{c|cccc}
\hline
\rule{0pt}{2.6ex}
 $k=1,N=50$ & ADMM & RLS & LRR & HOLRR \\
\hline
 linear & $0.6323$ & $0.6481$ & $\underline{0.6261}$ & $0.6310$ \\
 rbf &  - &0.6081 & 0.6076 & \textbf{\underline{0.6035}} \\
 poly &  - &0.6580 & \underline{0.6526} & 0.6793\vspace{0.1cm}\\
 \hline
\rule{0pt}{2.6ex}
$k=1,N=100$ & ADMM & RLS & LRR & HOLRR \\
\hline
linear & \underline{0.5997} & 0.6218 & \underline{0.6003} & \underline{0.6003} \\
rbf &  - &0.5926 & 0.5920 & \textbf{\underline{0.5808}} \\
poly &  - &0.6230 & \underline{0.6160} & 0.6222\vspace{0.1cm}\\
\hline
\rule{0pt}{2.6ex}
 $k=3,n=50$ & ADMM & RLS & LRR & HOLRR \\
 \hline
 linear & 0.6367 & 0.6683 & 0.6256 & \underline{0.6163} \\
 rbf &  - &0.6297 & 0.6296 & \textbf{\underline{0.6096}} \\
 poly &  - &0.6587 & 0.6412 & \underline{0.6355}\vspace{0.1cm}\\
\hline
\rule{0pt}{2.6ex}
  $k=3,N=100$ & ADMM & RLS & LRR & HOLRR \\
  \hline
  linear & 0.6269 & 0.6740 & 0.6186 & \underline{0.6086} \\
  rbf &  - &0.6153 & 0.6090 & \textbf{\underline{0.6015}} \\
  poly &  - &0.6588 & 0.6439 & \underline{0.6379}\vspace{0.1cm}\\
\hline
\rule{0pt}{2.6ex}  
$k=5,N=50$   & ADMM & RLS & LRR & HOLRR \\
\hline
linear & 0.6302 & 0.6719 & 0.6232 & \underline{0.6127} \\
rbf &  - &0.6150 & 0.6192 & \textbf{\underline{0.6017}} \\
poly &  - &0.6579 & \underline{0.6338} & 0.6353\vspace{0.1cm}\\
\hline
\rule{0pt}{2.6ex}
 $k=5,N=100$ & ADMM & RLS & LRR & HOLRR \\
 \hline
 linear & 0.6140 & 0.6449 & 0.6021 & \underline{0.5971} \\
 rbf &  - &0.6020 & 0.5908 & \textbf{\underline{0.5886}} \\
 poly &  - &0.6432 & 0.6202 & \underline{0.6107}\\
 
\hline
\end{tabular}
 
  \end{sc}
 \end{scriptsize}
 
 \end{center}

\vspace{-0.5cm}

  \end{table}

\section{Conclusion}
\label{sec:ccl}
We proposed a low-rank multilinear regression model for tensor-structured output data.
We developed a fast and efficient algorithm to tackle the multilinear rank penalized
minimization problem and provided theoretical approximation guarantees. Experimental
results showed that capturing low-rank structure in the output data can help to 
improve tensor regression performance.

\newpage

\newpage
\bibliographystyle{icml2016}
\bibliography{bib_TensorKernels}

\end{document}


\twocolumn[
\icmltitle{Higher-Order Low-Rank Regression (Supplementary Material)}

\icmlauthor{Your Name}{email@yourdomain.edu}
\icmladdress{Your Fantastic Institute,
            314159 Pi St., Palo Alto, CA 94306 USA}
\icmlauthor{Your CoAuthor's Name}{email@coauthordomain.edu}
\icmladdress{Their Fantastic Institute,
            27182 Exp St., Toronto, ON M6H 2T1 CANADA}

\icmlkeywords{boring formatting information, machine learning, ICML}

\vskip 0.3in

Here goes the proofs. Here goes the proofs. Here goes the proofs. Here goes the proofs. Here goes the proofs. Here goes the proofs. Here goes the proofs. Here goes the proofs. Here goes the proofs. Here goes the proofs. Here goes the proofs. Here goes the proofs. Here goes the proofs. Here goes the proofs. Here goes the proofs. Here goes the proofs. Here goes the proofs. Here goes the proofs. Here goes the proofs. Here goes the proofs. Here goes the proofs. Here goes the proofs. Here goes the proofs. 

\bibliographystyle{icml2016}
\bibliography{bib_TensorKernels}

]